\documentclass{article} 

\usepackage[final]{nips_2016_nomarkings}

\usepackage{hyperref}
\usepackage{url}
\usepackage{subcaption}

\usepackage{amsfonts}  
\usepackage{graphicx}
\usepackage{caption}
\usepackage{amsmath}

\usepackage{wrapfig}
\usepackage{amssymb,amsmath,amsthm,mathtools}

\usepackage{appendix}
\usepackage{etoolbox}
\usepackage{lipsum}
\usepackage{natbib}
\usepackage{booktabs}

\usepackage{xr}

\newcommand{\sectionname}[0]{Section~}
\renewcommand{\figurename}[0]{Figure~}

\title{Understanding intermediate layers \\
       using linear classifier probes}

\author{Guillaume Alain \\
Mila, University of Montreal \\
guillaume.alain.umontreal@gmail.com
\and
Yoshua Bengio \\
Mila, University of Montreal}

\newcommand{\paperroot}{.}

\begin{document}

\maketitle

\begin{abstract}
Neural network models have a reputation for being black boxes.
We propose to monitor the features at every layer of a model
and measure how suitable they are for classification.
We use linear classifiers, which we refer to as ``probes'',
trained entirely independently of the model itself.

This helps us better understand the roles and dynamics
of the intermediate layers.
We demonstrate how this can be used to develop a better intuition
about models and to diagnose potential problems.

We apply this technique to the popular models Inception v3
and Resnet-50. Among other things, we observe experimentally
that the linear separability of features increase monotonically
along the depth of the model.
\end{abstract}



\section{Introduction}
\label{lcp-sec:introduction}

The recent history of deep neural networks features an impressive
number of new methods and technological improvements to allow
the training of deeper and more powerful networks.

Deep neural networks still carry some of their original reputation
of being black boxes, but many efforts have been made
to understand better what they do, what is the role of each layer~\citep{yosinski2014transferable},
how we can interpret them~\citep{zeiler2014visualizing}
and how we can fool them~\citep{biggio2013evasion, szegedy2013intriguing}.

In this paper, we take the features of each layer separately
and we fit a linear classifier to predict the original classes.
We refer to these linear classifiers as ``probes'' and we make
sure that we never influence the model itself by taking
measurements with probes. We suggest that the reader think of
those probes as thermometers used to measure the temperature
simultaneously at many different locations.

More broadly speaking, the core of the idea is that there
are interesting quantities that we can report based on
the features of many independent layers if we allow
the ``measuring instruments'' to have their own trainable parameters
(provided that they do not influence the model itself).

In the context of this paper, we are working with convolutional
neural networks on image classification tasks on the
MNIST and ImageNet~\citep{ILSVRC15} datasets. 
Naturally, we fit linear classifier probes to predict those
classes, but in general it is possible to monitor the
performance of the features on any other objective.

Our contributions in this paper are twofold.

Firstly, we introduce these ``probes'' as a general tool
to understand deep neural networks. We show how they can be
used to characterize different layers, to debug bad models,
or to get a sense of how the training is progressing in a well-behaved model.
While our proposed idea shares commonalities with~\citet{montavon2011kernel},
our analysis is very different.


Secondly, we observe that the measurements of the probes are surprizingly
monotonic, which means that the degree of linear separability
of the features of layers increases as we reach the deeper layers.
The level of regularity with which this happens is surprizing given
that this is not technically part of the training objective.
This helps to understand the dynamics of deep neural networks.


\section{Related Work}


Many researchers have come up with techniques to analyze certain aspects of
neural networks which may guide our intuition and provide a partial explanation
as to how they work.

In this section we will provide a survey of the literature on the
subject, with a little more focus on papers related our current work.

\subsection{Linear classification with kernel PCA}
\label{lcp-sec:kernel-PCA}

In our paper we investigate the linear separability of the features found at
intermediate layers of a deep neural network.

A similar starting point is presented by~\citet{montavon2011kernel}.
In that particular case, the authors use kernel PCA to project the features of
a given layer onto a new representation which will then be used to fit the best
linear classifier. They use a radial basis function as kernel, and they choose
to project the features of individual layers by using the $d$ leading eigenvectors
of the kernel PCA decomposition.
They investigate the effects that $d$ has on the quality of the linear classifier.

Naturally, for a sufficiently large $d$, it would be possible to overfit on the
training set (given how easy this is with a radial basis function),
so they consider the situation where $d$ is relatively small.
They demonstrate that, for deeper layers in a neural network,
they can achieve good performance with smaller $d$.
This suggests that the features of the original convolution neural network are
indeed more ``abstract'' as we go deeper, which corresponds to the
general intuition shared by many researchers.

They explore convolution networks of limited depth with a restricted subset of
10k training samples of MNIST and CIFAR-10.

\subsection{Generalization and transferability of layers}


There are good arguments to support the claim that
the first layers of a convolution network for image recognition
contain filters that are relatively ``general'', in the sense
that they would work great even if we switched to an entirely
different dataset of images. The last layers are specific
to the dataset being used, and have to be retrained when using
a different dataset.
In~\citet{yosinski2014transferable} the authors try to pinpoint the layer at which
this transition occurs, but they show that the exact transition
is spread across multiple layers.
In~\citet{donahue2014decaf} the authors study the transfer of features from
the last few layers of a model to a novel generic task.
In~\citet{zeiler2014visualizing} the authors show that the filters
are picking up certain patterns that make sense to us visually, and they
show a method to visually inspect the filters as input images.

\subsection{Relevance Propagation}

In~\citet{bach2015pixel}, the authors introduce the idea of \emph{Relevance Propagation}
as a way to identify which pixels of the input space are the most important to
the classifier on the final layer.
Their approach frames the ``relevance'' as a kind of quantity that is to be
preserved across the layers, as a sort of shared responsibility to be divided among
the features of a given layer.

In~\citet{binder2016layer} the authors apply the concept of Relevance Propagation
to a larger family of models. Among other things, they provide a nice experiment
where they study the effects of corrupting the pixels deemed the most relevant,
and they show how this affects performance more than corrupting randomly-selected pixels
(see Figure 2 of their paper). See also~\citet{lapuschkin2016analyzing}.
Other research dealing with Relevance Propagation includes~\citet{arras2017explaining}
where this is applied to RNN in text.

We would also note that a good number of papers on interpretability of
neural networks deals with ``interpretations'' taking the form of regions of the
original image being identified, or where the pixels in the original image receive
a certain value of how relevant they are (e.g. a heat map of relevance).

In those cases we rely on the human user to parse the regions of the image with
their vision so as to determine whether the region indeed makes sense or whether
the information contained within is irrelevant to the task at hand.
This is analogous to the way that image-captioning attention~\citep{xu2015show}
can highlight portions of the input image that inspired specific segments of the caption.

An interesting approach is presented in~\citet{mahendran2015understanding, mahendran2016visualizing, dosovitskiy2016inverting}
where the authors analyze the set of ``equivalent'' inputs in the sense that
some of the features at a given layer should be preserved.
Given a layer to study, they apply a regularizer (e.g. total variation)
and use gradient descent in order to reconstruct the pre-image that yields the
same features at that layer, but for which the regularizer would be minimized.
This procedure yields pre-images that are of the same format as the input image,
and which can be used to get a sense of what are the components of the original
image that are preserved. For certain tasks, one may be surprised as to how many
details of the input image are being completely discarded by the time we reach
the fully-connected layers at the end of a convolution neural network.

\subsection{SVCCA}


In~\citet{raghu2017bottom,raghu2017svcca} the authors study the question of whether neural networks are trained from the first to the last layer, or the other way around (i.e. ``bottom up'' vs ``top down''). The concept is rather intuitive, but it still requires a proper definition of what they mean. They use Canonical Correlation Analysis (CCA) to compare two instances of a given model trained separately. Given that two different instances of the same model might assign entirely different roles to their neurons (on corresponding layers), this is a comparison that is normally impossible to even attempt. 

On one side, they take a model that has already been optimized. On the other side, they take multiple snapshots of a model during training. Every layer of one model is being compared with every other layer of the other. The values computed by CCA allows them to report the correlation between every pair of layers. This shows how quickly a given layer of the model being trained is going to achieve a configuration equivalent to the one of the optimized model. They find that the early layers reach their final configuration, so to speak, much earlier than layers downstream.

Given that any two sets of features can be compared using CCA, they also compare the correlation between any intermediate layer and the ground truth. This gives a sense of how easy it would be to predict the target label using the features of any intermediate layer instead of only using the last layer (as convnet usually do). Refer to Figure 6 of~\citet{raghu2017svcca} for more details. This aspect of~\citet{raghu2017svcca} is very similar to our own previous work~\citep{alain2016understanding}.






\section{Monitoring with probes}

\subsection{Information theory, and monotonic improvements to linear separability}
\label{lcp-sec:information-theory-and-monotonic-decrease}

The initial motivation for linear classifier probes was related to a reflection
about the nature of information (in the entropy sense of the word) passing
from one layer to the next.

New information is never added as we propagate forward in a model.
If we consider the typical image classification problem,
the representation of the data is transformed over the course
of many layers, to be finally used by a linear classifier at the last layer.

\begin{figure}[h]
\center
\begin{subfigure}[h]{0.45\textwidth}
  \includegraphics[trim = 10mm 10mm 10mm 10mm, clip, width=\linewidth]{\paperroot/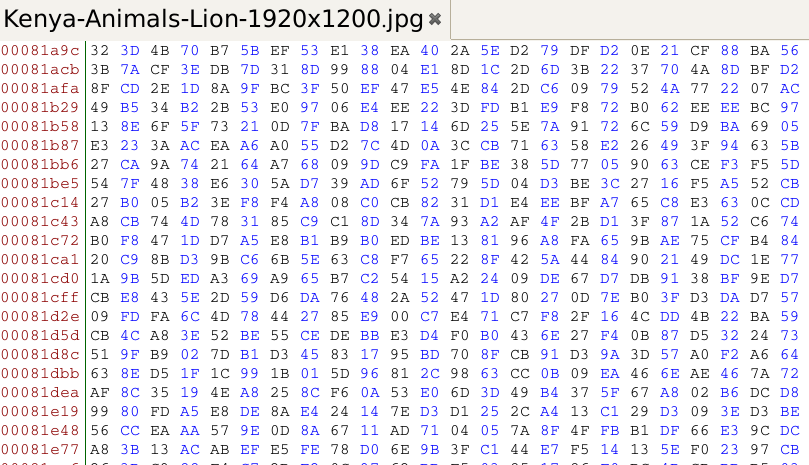}
  \caption{hex dump of picture of a lion}
  \label{lcp-fig:experiment_019_prediction_error_00000}
\end{subfigure}
\hfill
\begin{subfigure}[h]{0.45\textwidth}
  \includegraphics[trim = 10mm 10mm 10mm 10mm, clip, width=\linewidth]{\paperroot/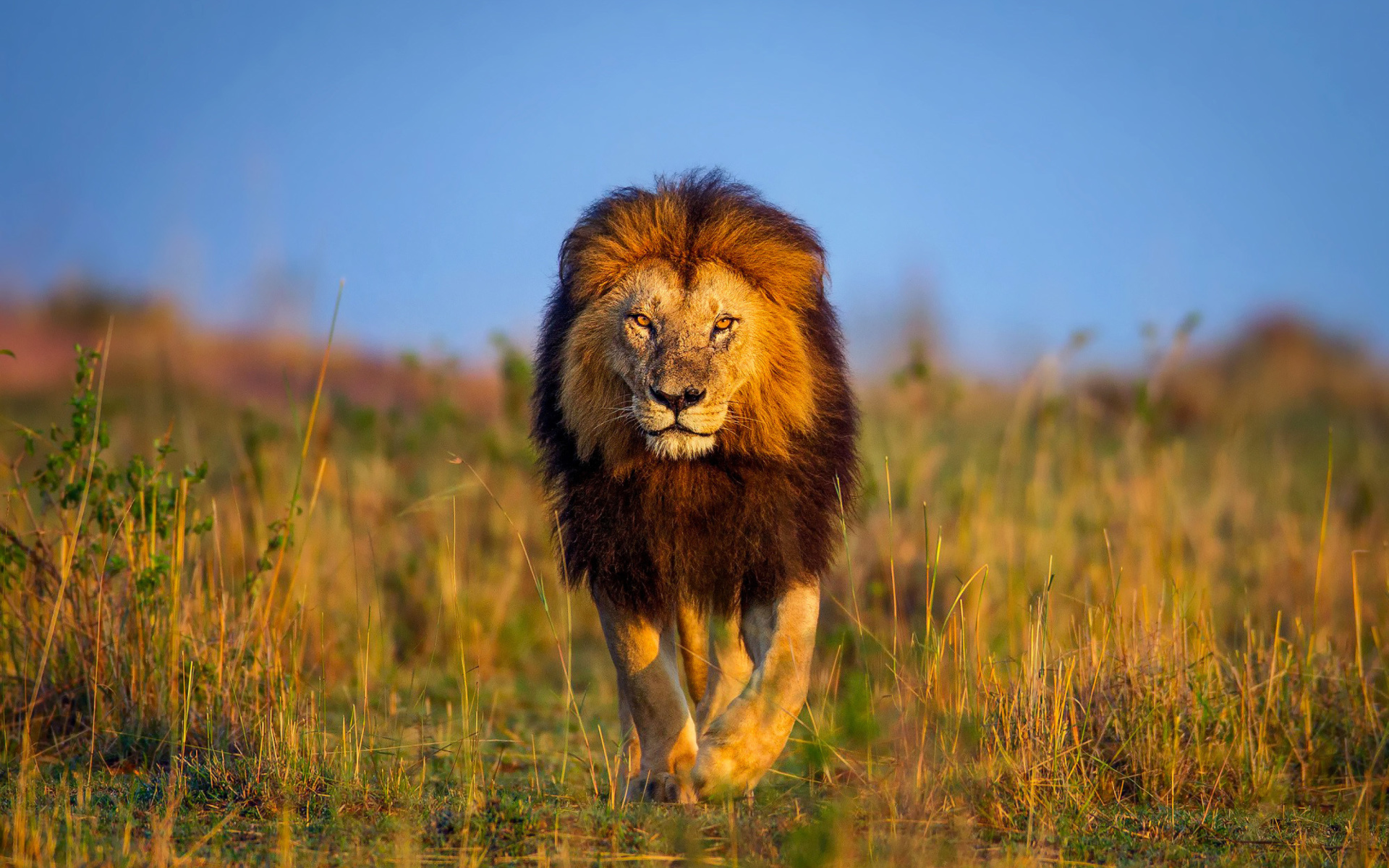}
  \caption{same lion in human-readable format}
  \label{lcp-fig:experiment_019_prediction_error_02990}
\end{subfigure}
\caption[Lion picture versus hex dump]{The hex dump represented at the left has more information contents
         than the image at the right. Only one of them can be processed
         by the human brain in time to save their lives.
         Computational convenience matters. Not just entropy.
        }
\label{lcp-fig:lions}
\end{figure}

In the case of a binary classifier (say, detecting the presence or absence of a lion
in a picture of the savannah like in \figurename \ref{lcp-fig:lions}),
we could say that there was at most one bit of information to
be uncovered in the original image. Lion or no lion ?
Here we are not interested in measuring the information about the
pixels of an image that we want to reconstruct. That would be a different problem.

This is illustrated in a formal way by the \emph{Data Processing Inequality}.
It states that, for a set of three random variables satisfying the
dependency
\begin{equation*}
X \rightarrow Y \rightarrow Z
\end{equation*}
then we have that
\begin{equation*}
I(X;Z) \leq I(X; Y)
\end{equation*}
where $I(X,Y)$ is the mutual information.

The task of a deep neural network classifier is to come up with a representation for the
final layer that can be easily fed to a linear classifier (i.e. the most elementary
form of useful classifier). The cross-entropy loss applies a lot of pressure
directly on the last layer to make it linearly separable.
Any degree of linear separability in the intermediate layers happens only as a by-product. 


On one hand, we have that every layer has less \emph{information} than its parent layer. On the other hand, we observe experimentally in \sectionname \ref{lcp-sec:mnist}, \ref{lcp-sec:resnet50} and \ref{lcp-sec:inceptionv3} that features from deeper layers work better with linear classifiers to predict the target labels. At first glance this might seem like a contradiction.

One of the important lessons is that neural networks are really about distilling computationally-useful \emph{representations}, and they are not about \emph{information contents} as described by the field of Information Theory.

\subsection{Linear classifier probes}
\label{lcp-sec:linear-classifier-probes-definition}

Consider the common scenario in deep learning
in which we are trying to classify the input data $X$
to produce an output distribution over $D$ classes.
The last layer of the model is a densely-connected map
to $D$ values followed by a softmax, and we train by
minimizing cross-entropy.


At every layer we can take the features $H_k$ from that layer
and try to predict the correct labels $y$ using a linear classifier
parameterized as
\begin{align*}
  f_k \colon H_k &\to [0, 1]^D\\
  h_k &\mapsto \textrm{softmax}\left( Wh_k + b\right).
\end{align*}
where $h_k \in H$ are the features of hidden layer $k$,
$[0, 1]^D$ is the space of categorical distributions of the $D$ target classes,
and $(W, b)$ are the probe weights and biases to be learned
so as to minimize the usual cross-entropy loss.

Let $\mathcal{L}_k^\textrm{train}$ be the empirical loss
of that linear classifier $f_k$ evaluated over the training set.
We can also define $\mathcal{L}_k^\textrm{valid}$ and $\mathcal{L}_k^\textrm{test}$
by exporting the same linear classifier on the validation and test sets.

Without making any assumptions about the model itself being trained,
we can nevertheless assume that these $f_k$ are themselves optimized
so that, at any given time, they reflect the currently optimal thing
that can be done with the features present.

We refer to those linear classifiers as ``probes'' in an effort
to clarify our thinking about the model. These probes do not affect
the model training. They only measure the level of linear separability
of the features at a given layer. Blocking the backpropagation from
the probes to the model itself can be achieved by using \texttt{tf.stop\_gradient}
in Tensorflow (or its Theano equivalent), or by managing the probe parameters
separately from the model parameters.


Note that we can avoid the issue of local minima because training
a linear classifier using softmax cross-entropy is a convex problem.

\vspace{1em}

In this paper, we study
\begin{itemize}
 \item how $\mathcal{L}_k$ decreases as $k$ increases (see \sectionname \ref{lcp-sec:information-theory-and-monotonic-decrease}),
 \item the usefulness of $\mathcal{L}_k$ as a diagnostic tool (see \sectionname \ref{lcp-sec:patho-skip-connections}).
\end{itemize}

\subsection{Practical concern : $\mathcal{L}_k^\textrm{train}$ vs $\mathcal{L}_k^\textrm{valid}$}
\label{lcp-sec:practical-concern-train-vs-valid}

The reason why we care about optimality of the probes in \sectionname \ref{lcp-sec:linear-classifier-probes-definition}
is because it abstracts away the problem of optimizing them.
When a general function $g(x)$ has a unique global minimum,
we can talk about that minimum without ambiguity even though,
in practice, we are probably going to use only a convenient
approximation of the minimum.

This is acceptable in a context where we are seeking
better intuition about deep learning models by using linear classifier probes.
If a researcher judges that the measurements are useful to further
their understanding of their model (and act on that intuition),
then they should not worry too much about how close they are to optimality.

This applies also to the question of whether we should prioritize
$\mathcal{L}_k^\textrm{train}$ or $\mathcal{L}_k^\textrm{valid}$.
We would argue that $\mathcal{L}_k^\textrm{valid}$ seems like a
more meaningful quantity to monitor, but depending on our experimental setup it might
not be easy to track $\mathcal{L}_k^\textrm{valid}$ in all circumstances.

Moreover, for the purposes of many of the experiments in this paper
we chose to report the classification error instead of the cross-entropy,
since this is ultimately often the quantity that matters the most.
Reporting the top5 classification error could also have been possible.

\subsection{Practical concern : Dimension reduction on features}
\label{lcp-sec:dropping-features}

Another practical problem can arise when certain layers of a neural network
have an exceedingly large quantity of features.
The first few layers of Inception v3, for example,
have a few million features when we multiply height, width and channels.
This leads to parameters for a single probe taking upwards of a few gigabytes of
storage, which is disproportionately large when we consider that
the entire set of model parameters takes less space than that.

In those cases, we have three possible suggestions for trimming down the
space of features on which we fit the probes.
\begin{itemize}
 \item Use only a random subset of the features (but always the same ones).
       This is used on the Inception v3 model in \sectionname \ref{lcp-sec:inceptionv3}.
 \item Project the features to a lower-dimensional space. Learn this mapping.
       This is probably a worse idea than it sounds because the projection matrix itself
       can take a lot of storage (even more than the probe parameters).
 \item When dealing with features in the form of images (height, width, channels),
       we can perform 2D pooling along the (height, width) of each channel.
       This reduces the number of features to the number of channels.
       This is used on the ResNet-50 model in \sectionname \ref{lcp-sec:resnet50}.
\end{itemize}

In practice, when using linear classifier probes on any serious model (i.e. not MNIST)
we have to choose a way to reduce the number of features used.

Note that we also want to avoid a situation where our probes
are simply overfitting on the features because there are too many features.
It was recently demonstrated that very large models can fit random labels
on ImageNet~\citep{zhang2016understanding}. This is a situation
that we want to avoid because the probe measurements would be entirely meaningless
in that situation. Dimensionality reduction helps with this concern.

\subsection{Basic example on MNIST}
\label{lcp-sec:mnist}

In this section we run the MNIST convolutional model provided by the
\texttt{tensorflow/models} github repository (\texttt{image/mnist/convolutional.py}).
We selected that model
for reproducibility and to demonstrate how to easily peek into popular models
by using probes.

\begin{figure}[ht]
\center
  \includegraphics[width=\linewidth]{\paperroot/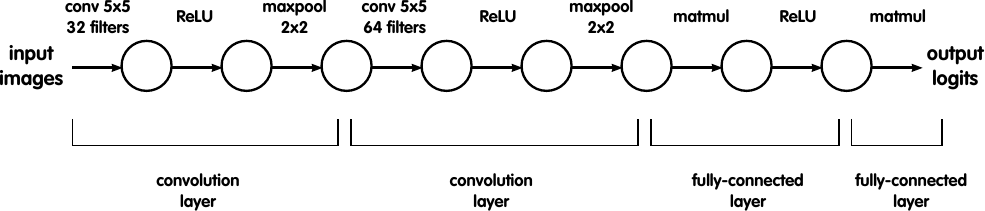}
  \caption[Elementary convnet with inserted probes]{This graphical model represents the neural network that we are going to use for MNIST.
           The model could be written in a more compact form, but we represent it
           this way to expose all the locations where we are going to insert probes.
           The model itself is simply two convolutional layers followed by two
           fully-connected layer (one being the final classifier).
           However, we insert probes on each side of each convolution,
           activation function, and pooling function.
           This is a bit overzealous, but the small size of the model makes this
           relatively easy to do.
  }
  \label{lcp-fig:mnist-model}
\end{figure}

\begin{figure}[ht]
\center
\begin{subfigure}[h]{0.45\linewidth}
  \includegraphics[width=\linewidth]{\paperroot/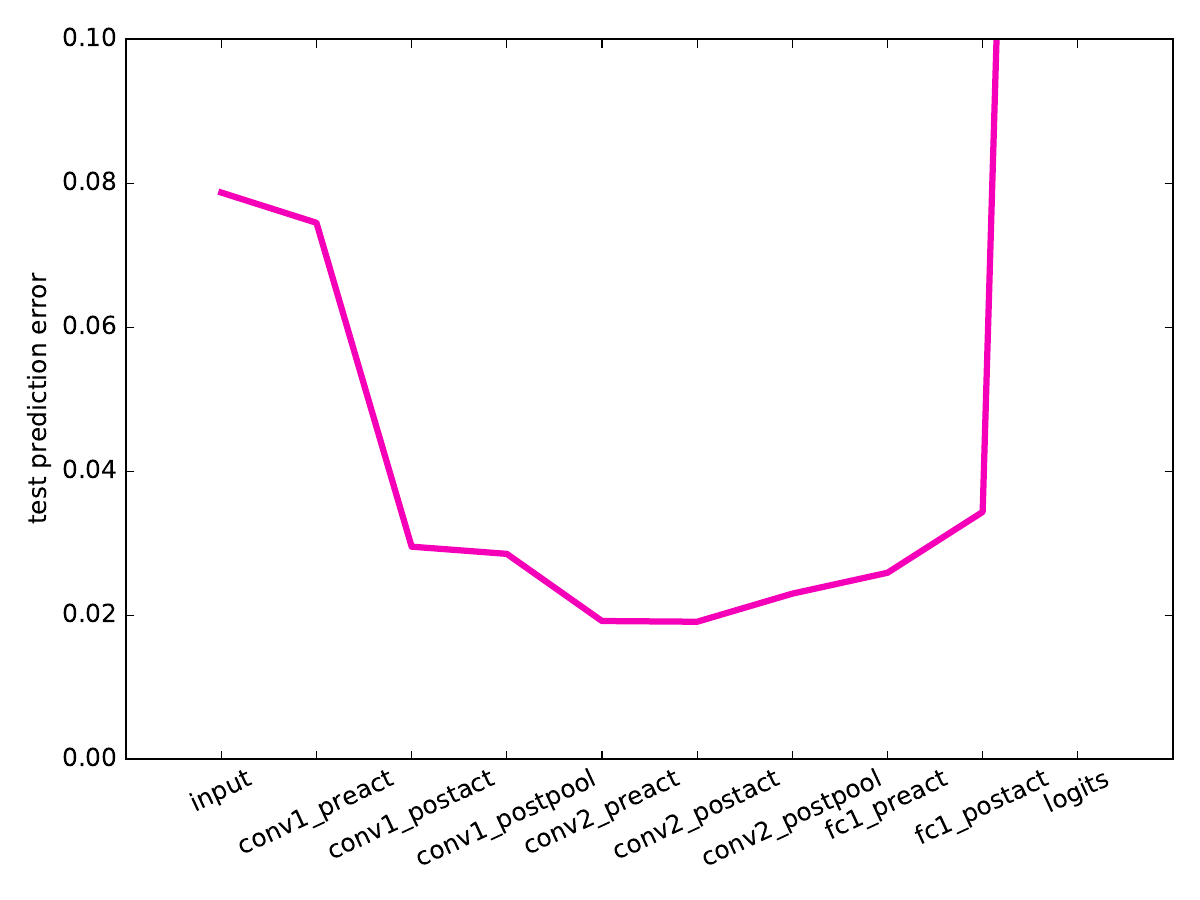}
  \caption{After initialization, no training.}
  \label{lcp-fig:mnist-init}
\end{subfigure}
\begin{subfigure}[h]{0.45\linewidth}
  \includegraphics[width=\linewidth]{\paperroot/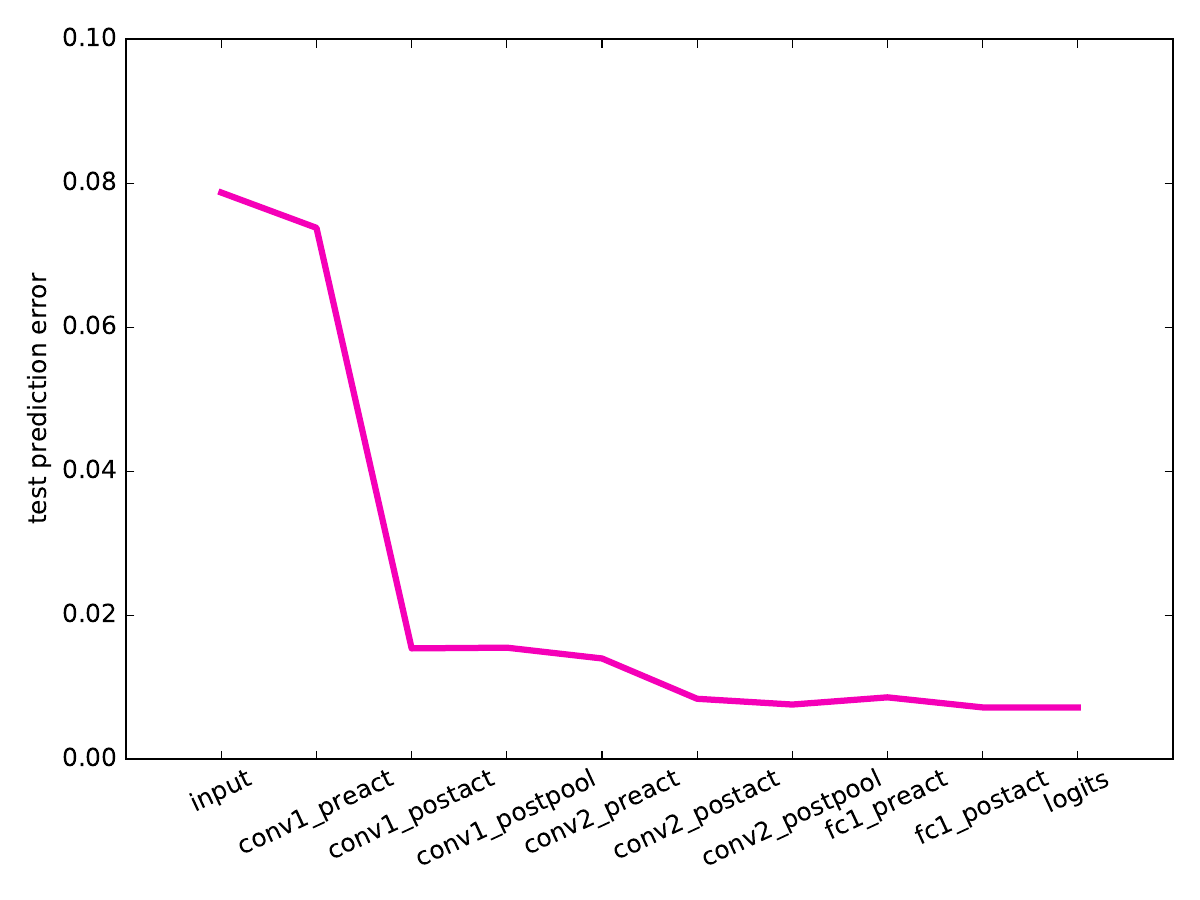}
  \caption{After training for 10 epochs.}
  \label{lcp-fig:mnist-after-10-epochs}
\end{subfigure}
\caption[Test error on MNIST for each probes]{We represent here the test prediction error for each probe,
        at the beginning and at the end of training.
        This measurement was obtained through early stopping based on
        a validation set of $10^4$ elements. The probes are prevented from overfitting
        the training data.
        We can see that, at the beginning of training (on the left), the randomly-initialized
        layers were still providing useful transformations.
        The test prediction error goes from 8\% to 2\% simply using those
        random features. The biggest impact comes from the first ReLU.
        At the end of training (on the right), the test prediction error
        is improving at every layer (with the exception of a minor kink on \texttt{fc1\_preact}).}
\label{lcp-fig:mnist}
\end{figure}

We start by sketching the model in \figurename~\ref{lcp-fig:mnist-model}.
We report the results at the beginning and the end of training on \figurename~\ref{lcp-fig:mnist}.
One of the interesting dynamics to be observed there is how useful the first layers are,
despite the fact that the model is completely untrained.
Random projections can be useful to classify data, and this has
been studied by others~\citep{jarrett2009best}.


\subsection{Other objectives}
\label{lcp-sec:other-objectives}

Note that it would be entirely possible to use linear classifier probes
on a different set of labels. For the same reason as it is possible to
transfer many layers from one vision task to another (e.g. with different classes),
we are not limited to fitting probes using the same domain.

Inserting probes at many different layers of a model is essentially a way to ask the
following question:
\begin{center}
\emph{Is there any information about factor \hspace{0.1em} \textunderscore \textunderscore \textunderscore \textunderscore \textunderscore \textunderscore \hspace{0.15em} present  in this part of the model ?}
\end{center}

\section{Experiments with popular models}

\subsection{ResNet-50}
\label{lcp-sec:resnet50}

The family of ResNet models~\citep{he2016deep} are characterized by their large quantities of
\emph{residual layers} mapping essentially \hbox{$x \mapsto x + r(x)$}.
They have been very successful and there are various papers
seeking to understand better how they work~\citep{veit2016residual, larsson2016fractalnet, singh2016swapout}.

Here we are going to show how linear classifier probes might be able
to help us a little to shed some light into the \hbox{ResNet-50} model.
We used the pretrained model from the github repo (\texttt{fchollet/deep-learning-models})
of the author of Keras~\citep{chollet2015keras}.

One of the questions that comes up when discussing ResNet models
is whether the successive layers are essentially performing the same
operation over many times, refining the representation just a little more
each time, or whether there is a more fundamental change of representation
happening.

\begin{figure*}[!ht]
\centering
\begin{minipage}[b]{0.30\linewidth}
\begin{tabular}{r|cc}
    \toprule
        &            &   probe valid      \\
      layer  &     topology       &    prediction      \\
      name   &    &    error \\
    \midrule
	      &               &             \\
    input\_1 & (224, 224, 3) & 0.99 \\
	      &               &             \\
    add\_1 & (28, 28, 256) &    0.94 \\
    add\_2 & (28, 28, 256) &    0.89 \\
    add\_3 & (28, 28, 256) &    0.88 \\
	      &                        &   \\
    add\_4 & (28, 28, 512) &    0.87 \\
    add\_5 & (28, 28, 512) &    0.82 \\
    add\_6 & (28, 28, 512) &    0.79 \\
    add\_7 & (28, 28, 512) &    0.76 \\
	      &                       &     \\
    add\_8 & (14, 14, 1024) &    0.77 \\
    add\_9 & (14, 14, 1024) &    0.69 \\
    add\_10 & (14, 14, 1024) &    0.67 \\
    add\_11 & (14, 14, 1024) &    0.62 \\
    add\_12 & (14, 14, 1024) &    0.57 \\
    add\_13 & (14, 14, 1024) &    0.51 \\
	    &                        &      \\
    add\_14 & (7, 7, 2048) &    0.41 \\
    add\_15 & (7, 7, 2048) &    0.39 \\
    add\_16 & (7, 7, 2048) &    0.31 \\
	      &                  &   \\
    \bottomrule
    \end{tabular}
\subcaption{Validation errors for probes. Comparing different layers. \hbox{Pre-trained} ResNet-50 on ImageNet dataset.}
\label{table:probe-valid-prediction-error}
\end{minipage}
\hfil
\begin{minipage}[b]{0.65\linewidth}
\begin{flushright}
\includegraphics[trim = 1mm 1mm 1mm 1mm, clip, width=0.80\linewidth]{\paperroot/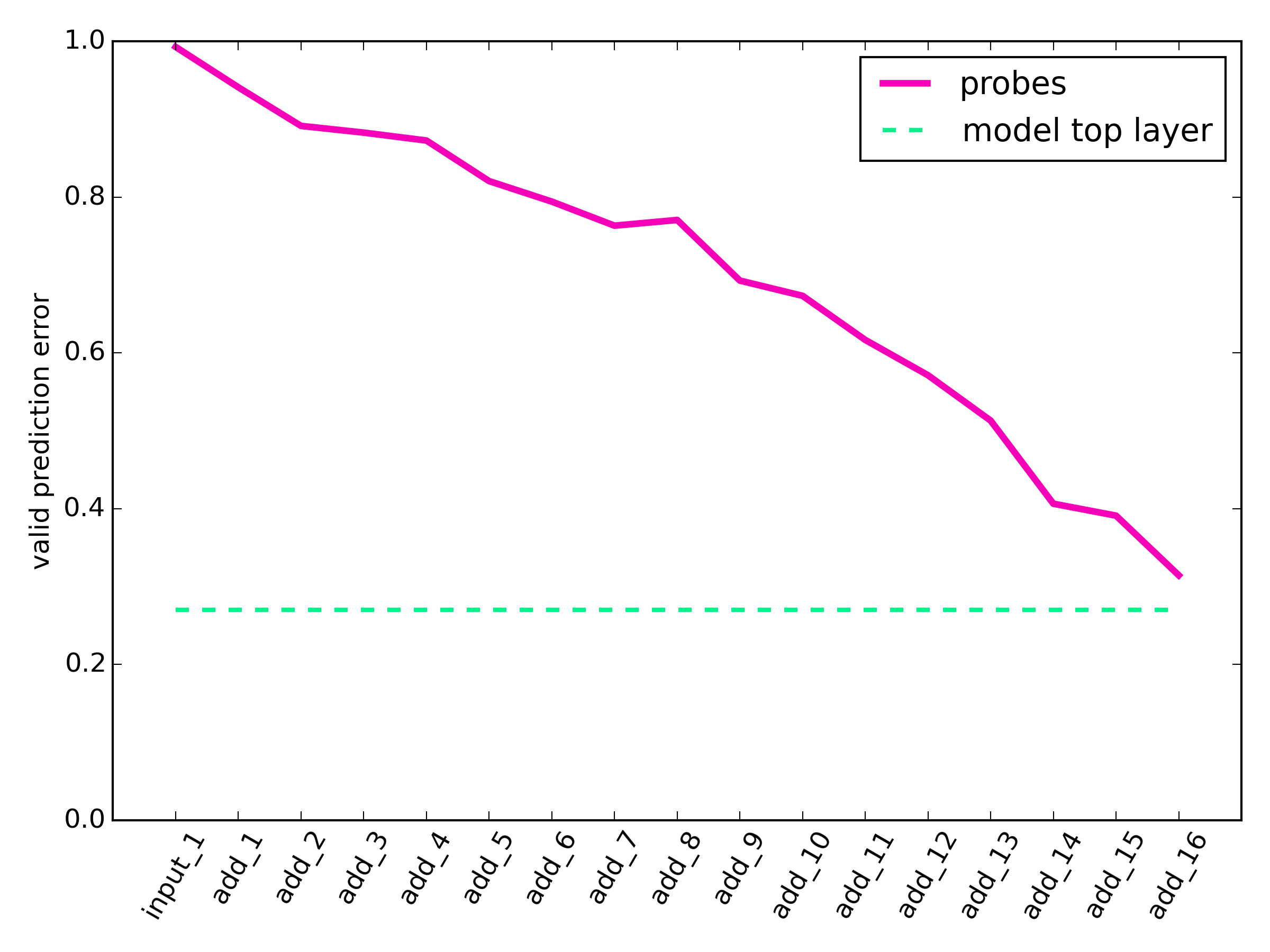}
\subcaption{
    Inserting probes at meaningful layers of ResNet-50.
    This plot shows the rightmost column of the table in \mbox{\figurename~\ref{table:probe-valid-prediction-error}}.
    Reporting the validation error for probes (magenta) and
    comparing it with the validation error of the pre-trained model (green).}
\end{flushright}                  
\label{lcp-fig:probe-valid-prediction-error}
\end{minipage}
\caption[Probes on ResNet-50 with ImageNet]{For the ResNet-50 model trained on ImageNet, we can see deeper features are better at predicting the output classes. More importantly, the relationship between depth and validation prediction error is almost perfectly monotonic. This suggests a certain ``greedy'' aspect of the representations used in deep neural networks. This property is something that comes naturally as a result of conventional training, and it is not due to the insertion of probes in the model.}
\label{lcp-fig:09423knsajf}
\end{figure*}

In particular, we can point to certain places in ResNet-50 where the image size
diminishes and we increase the number of channels. This happens at three
places in the model (identified with blank lines in Table \ref{table:probe-valid-prediction-error}).

\subsection{Inception v3}
\label{lcp-sec:inceptionv3}

\begin{figure*}[htb]
\centering
  \includegraphics[width=0.8\linewidth]{\paperroot/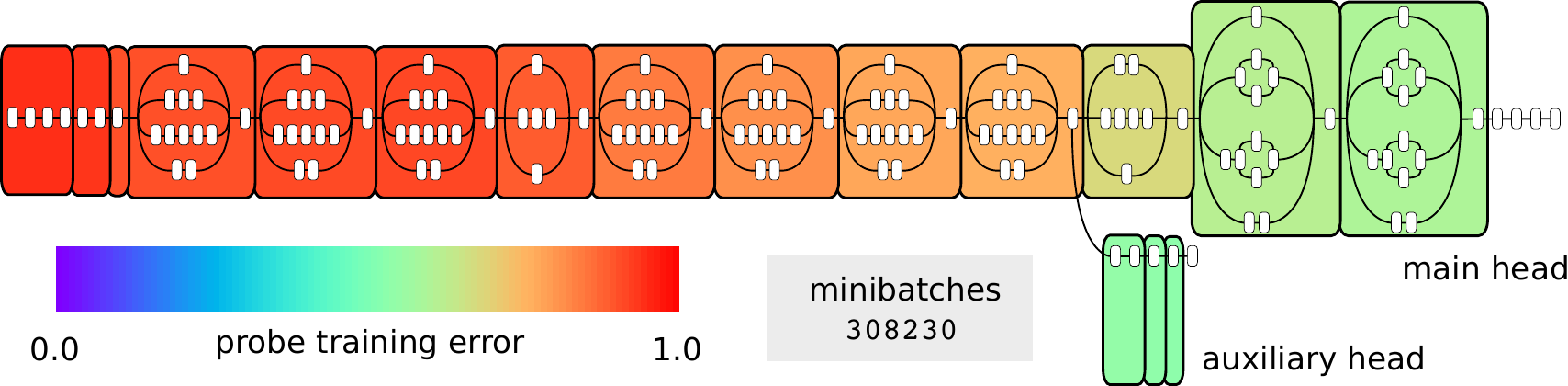}
      \caption[Inception v3 model after 2 weeks of training]{ Inception v3 model after 2 weeks of training.
            Red is bad (high prediction error) and green/blue is good (low prediction error).
            The smooth color gradient shows a very gradual transition in the degree of linear separability
            (almost perfectly monotonic).}
  \label{lcp-fig:inception-model}
\end{figure*}

We have performed an experiment using the Inception v3 model on
the ImageNet dataset~\citep{szegedy2015going, ILSVRC15}.
We show using colors in \figurename~\ref{lcp-fig:inception-model} how
the predictive error of each layer can be measured using probes.
This can be computed at many different times of training,
but here we report only after minibatch 308230, which corresponds
to about 2 weeks of training.

This model has a few particularities, one of which is that
it features an auxiliary branch that contributes to training
the model (it can be discarded afterwards, but not necessarily).
We wanted to investigate whether this branch is ``leading training'',
in the sense that its classifier might have lower prediction error
than the main head for the first part of the training.

This is something that we confirmed by looking at the prediction errors
for the probes, but the difference was not very large.
The auxiliary branch was ahead of the main branch by just a little.

The smooth gradient of colors in \figurename~\ref{lcp-fig:inception-model}
shows how the linear separability increases monotonically as we probe
layers deeper into the network.


Refer to the Appendix \sectionname \ref{lcp-appsec:inceptionv3} for
a comparison at four different moments of training, and for
some more details about how we reduced the dimensionality of
the feature to make this more tractable.

\section{Diagnostics for failing models}

\subsection{Pathological behavior on skip connections}
\label{lcp-sec:patho-skip-connections}

In this section we show an example of a situation where
we can use probes to diagnose a training problem as it is
happening.

\begin{figure}[h]
\center
\hfill
\begin{subfigure}[h]{0.40\textwidth}
  \vspace{2em} 
  \includegraphics[width=\linewidth]{\paperroot/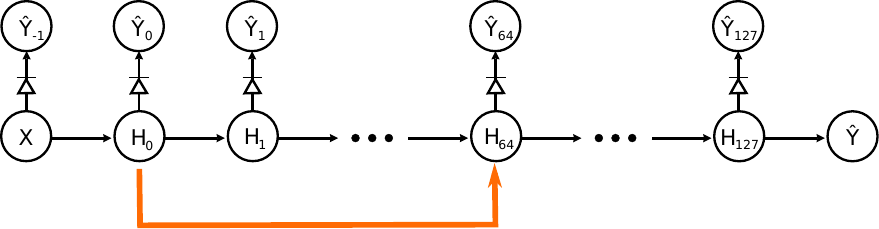}
  \caption{Model with 128 layers. A skip connection goes
           from the beginning straight to the middle of the graph.}
  \label{lcp-fig:graph-128bridge64}
\end{subfigure}
\hfill
\begin{subfigure}[h]{0.25\textwidth}
  \includegraphics[trim = 10mm 10mm 10mm 10mm, clip, width=\linewidth]{\paperroot/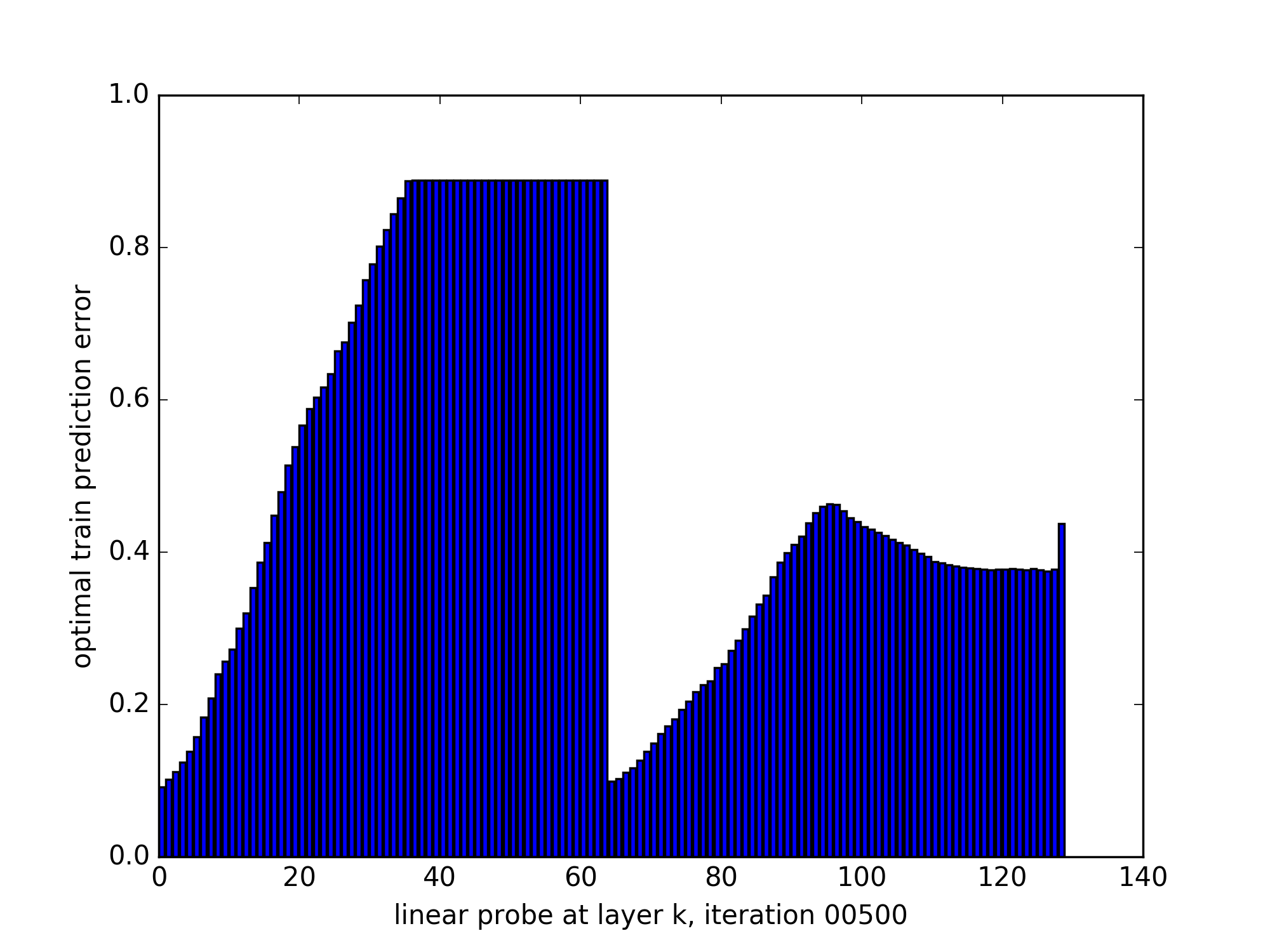}
  \caption{probes after 500 minibatches}
  \label{lcp-fig:graph-improvements-bridge-1000}
\end{subfigure}
\hfill
\begin{subfigure}[h]{0.25\textwidth}
  \includegraphics[trim = 10mm 10mm 10mm 10mm, clip, width=\linewidth]{\paperroot/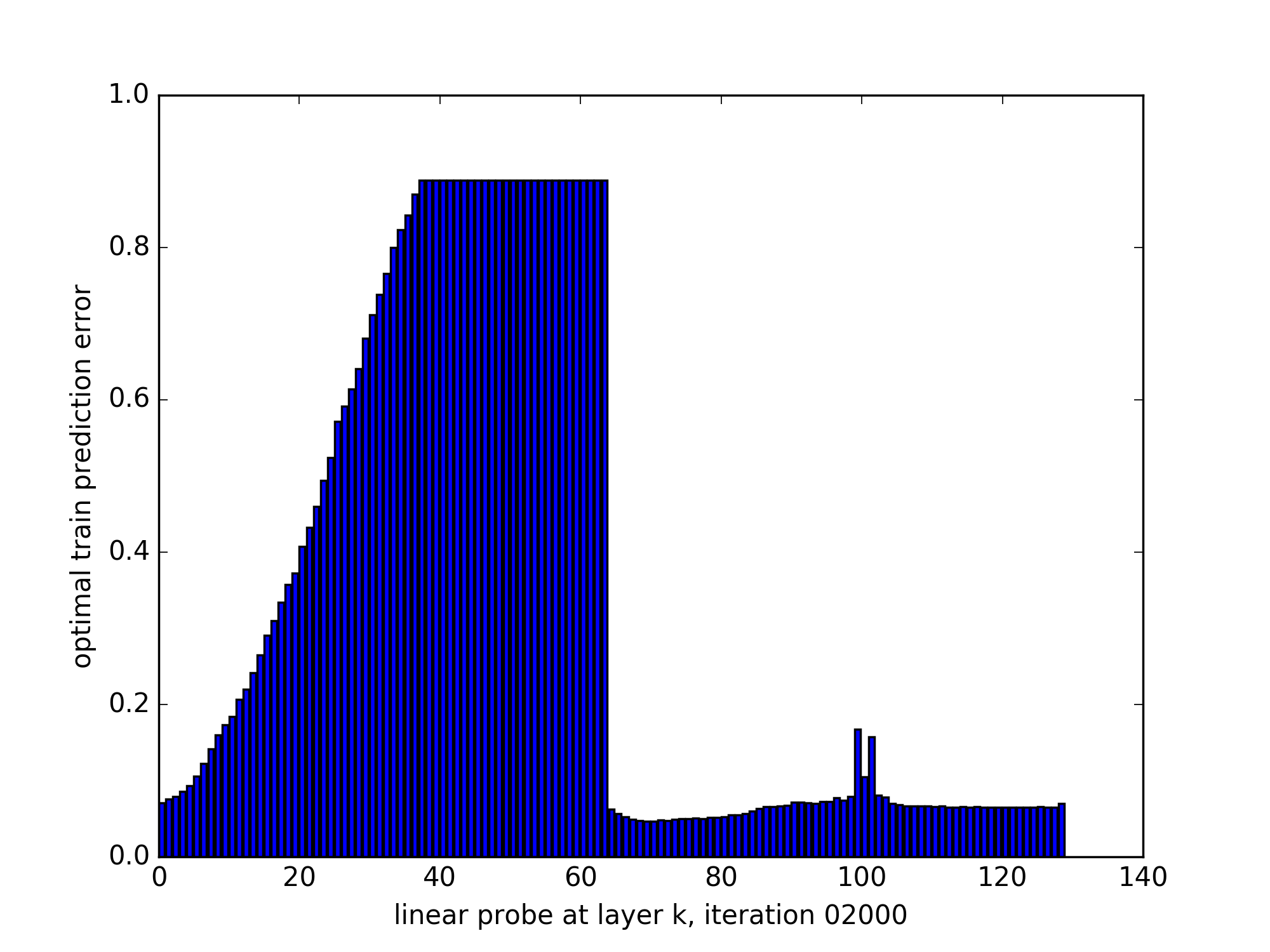}
  \caption{probes after 2000 minibatches}
  \label{lcp-fig:graph-improvements-bridge-2000}
\end{subfigure}
\caption[Pathological skip connection being diagnosed]{Pathological skip connection being diagnosed.
Refer to Appendix \sectionname \ref{lcp-appsec:diode-notation} for explanations about
the special notation for probes using the ``diode'' symbol.}
\label{lcp-fig:graph-improvements1}
\end{figure}

We purposefully selected a model that was
pathologically deep so that it would fail to train
under normal circumstances. We used 128 fully-connected layers
of 128 hidden units to classify MNIST, which is
not at all a model that we would recommend.
We thought that something interesting might happen
if we added a very long skip connection
that bypasses the first half of the model completely (\figurename~\ref{lcp-fig:graph-128bridge64}).

With that skip connection, the model became trainable
through the usual SGD. Intuitively, we thought that
the latter portion of the model would see use at first,
but then we did not know whether the first half of the
model would then also become useful.

Using probes we show that this solution was not working as intended,
because half of the model stays unused. The weights are not zero,
but there is no useful signal passing through that segment.
The skip connection left a dead segment and skipped over it.

The lesson that we want to show the reader is not that skip connections are bad.
Our goal here is to show that linear classification probes are a tool
to understand what is happening internally in such situations.
Sometimes the successful minimization of a loss fails to capture important details.

\section{Discussion and future work}
\label{lcp-sec:discussion-and-future-work}

We have presented a combination of both a small convnet on MNIST
and larger popular convnets Inception v3 and ResNet-50.
It would be nice to continue this work and look at ResNet-101, ResNet-151,
VGG-16 and VGG-19.
A similar thing could be done with popular RNNs also.

To apply linear classifier probes to a different context,
we could also try any setting where either Generative Adversarial Networks~\citep{goodfellow2014generative_alt}
or adversarial examples are used~\citep{szegedy2013intriguing}.

The idea of multi-layer probes has been suggested to us on multiple occasions. This could be seen as a natural extension
of the linear classifier probes.
One downside to this idea is that we lose the convexity property of the probes.
It might be worth pursuing in a particular setting,
but as of now we feel that it is premature to start using multi-layer probes.
This also leads to the convoluted idea of having a regular probe inside a multi-layer probe.


One completely new direction would be to train a model in a way that
actively discourages certain internal layers to be useful to linear classifiers.
What would be the consequences of this constraint?
Would it handicap a given model or would the model simply adjust without any trouble?
At that point, we are no longer dealing with non-invasive probes,
but we are feeding a strange kind of signal back to the model.

Finally, we think that it is rather interesting that the probe prediction errors
are almost perfectly monotonically decreasing. We suspect that this
warrants a deeper investigation into the reasons why that it happens,
and it may lead to the discovery of fundamental concepts to understand better
deep neural networks (in relation to their optimization). This is connected to the work done by~\citet{jastrzebski2017residual}.

\section{Conclusion}
\label{sec:conclusion}

In this paper we introduced the concept of the \emph{linear classifier probe}
as a conceptual tool to better understand the dynamics inside a neural network
and the role played by the individual intermediate layers.

We have observed experimentally that an interesting property holds :
the level of linear separability increases monotonically as we go to deeper layers.
This is purely an indirect consequence of enforcing this constraint on the last layer.

We have demonstrated how these probes can be used to identify
certain problematic behaviors in models that might not be apparent
when we traditionally have access to only the prediction loss and error.

We are now able to ask new questions and explore new areas.

We hope that the notions presented in this paper can contribute
to the understanding of deep neural networks and guide the intuition
of researchers that design them.

\subsubsection*{Acknowledgments}
Yoshua Bengio is a senior CIFAR Fellow. The authors would like to acknowledge the support of the
following agencies for research funding and computing support: NSERC, FQRNT, Calcul Qu\'ebec,
Compute Canada, the Canada Research Chairs and CIFAR. Thanks to Nicolas Ballas for fruitful
discussions, to Reyhane Askari and Mohammad Pezeshki for proofreading and comments, and to
all the reviewers for their comments.

\bibliography{alain17a}
\bibliographystyle{natbib}

\appendix


\section{Diode notation}
\label{lcp-appsec:diode-notation}

\begin{figure}[ht]
\center
\hfil
  \includegraphics[width=0.80\linewidth]{\paperroot/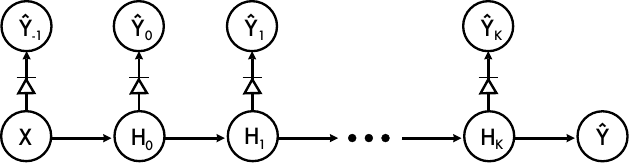}
\hfil
  \caption[Diode notation]{Probes being added to every layer of a model.
  These additional probes are not supposed to change the training of the model,
  so we add a little diode symbol through the arrows to indicate that the gradients will
  not backpropagate through those connections.}
  \label{appfig:probe_insertion}
\end{figure}

We have the following suggestion for extending traditional graphical models
to describe where probes are being inserted in a model. See \figurename ~\ref{appfig:probe_insertion}.

Due to the fact that probes do not contribute to backpropagation,
but they still consume the features during the feed-forward step,
we thought that borrowing the diode symbol from electrical engineering
might be a good idea. A diode is a one-way valve for electrical current.

This notation could be useful also outside of this context with probes,
whenever we want to sketch a graphical model and highlight the fact
that the gradient backpropagation signal is being blocked.

\section{Training probes with finished model}
\label{lcp-appsec:probe-training-after-the-fact}


Sometimes we do not care about measuring the probe losses/accuracy
during training, but we have a model that is already trained and
we want to report the measurements on that static model.

In that case, it is worth considering whether we really want to
augment the model by adding the probes and training the probes
by iterating through the training set. Sometimes the model itself
is computationally expensive to run and we can only do 150 images per second.
If we have to do multiple passes over the training set in order
to train probes, then it might be more efficient to run
the whole training set and extract the features to the local hard drive.
Experimentally, in the case for the pre-trained model \hbox{Resnet-50} (\sectionname \ref{lcp-sec:resnet50})
we found that we could process approximately
100 training samples per second when doing forward propagation, but we could
run through 6000 training samples per second when reading from the local hard drive.
This makes it a lot easier to do multiple passes over the training set.

%
%
%

\section{Inception v3}
\label{lcp-appsec:inceptionv3}

In \sectionname \ref{lcp-sec:dropping-features} we showed results
from an experiment using the Inception v3 model on the ImageNet dataset~\citep{szegedy2015going, ILSVRC15}. The results shown were
taken from the last training step only.

\begin{figure*}[ht]
\centering
  \includegraphics[trim = 0mm 0mm 0mm 0mm, clip, width=0.95\textwidth]{\paperroot/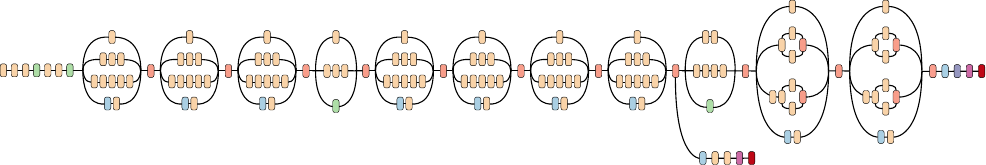}
  \caption[Sketch of Inception model]{Sketch of the Inception v3 model. Note the structure with the ``auxiliary head'' at the bottom, and the ``inception modules'' with a common topology represented as blocks that have 3 or 4 sub-branches.
  }
  \label{appfig:inception-model-sketch}
\end{figure*}

\begin{figure*}[ht]
\centering
  \includegraphics[width=0.8\linewidth]{\paperroot/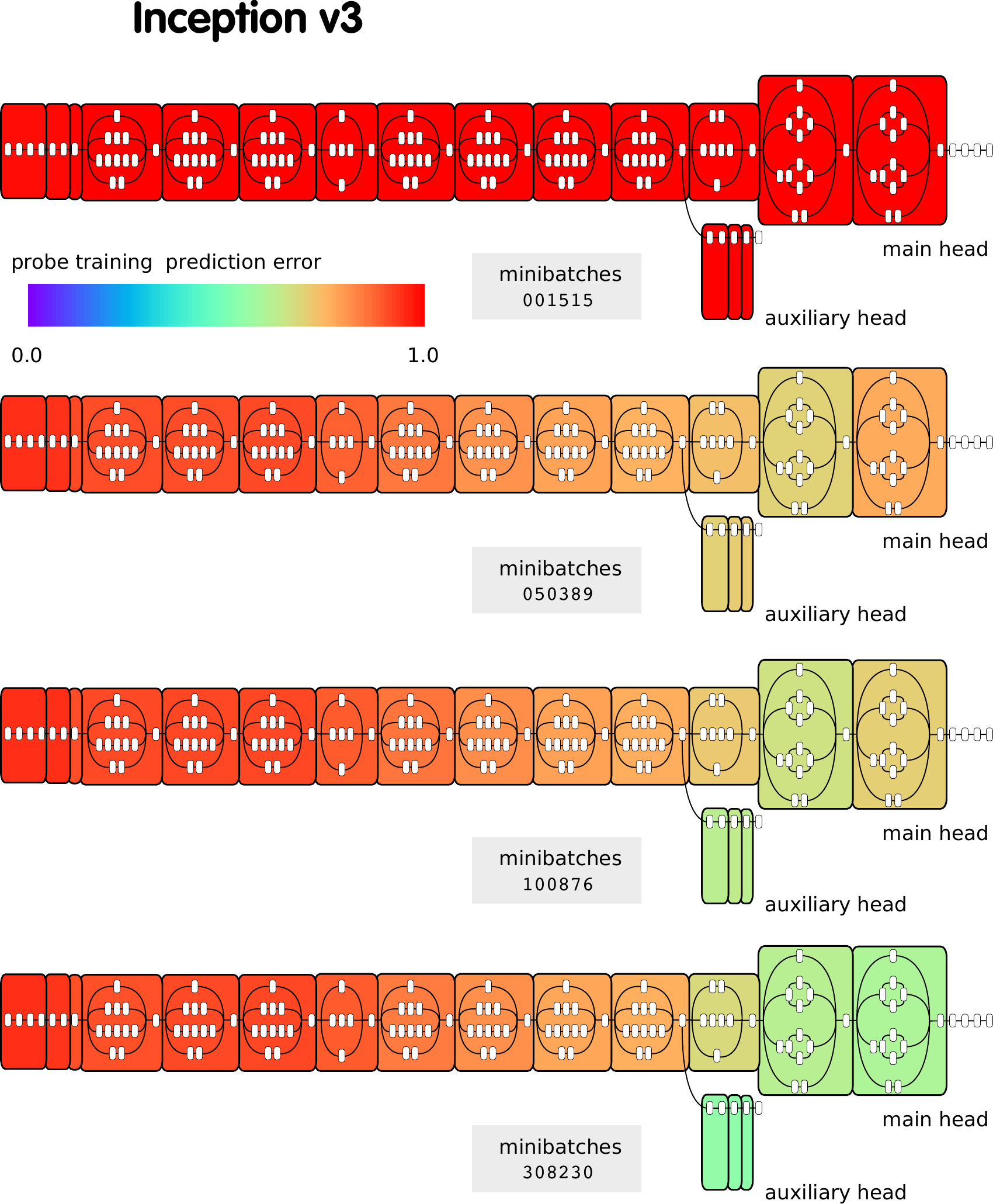}
  \caption[Probes on Inception at four checkpoints]{ Inserting a probe at multiple moments during training the Inception v3 model on the ImageNet dataset.
            We represent here the prediction error evaluated at a random subset of 1000 features.
            As expected, at first all the probes have a 100\% prediction error,
            but as training progresses we see that the model is getting better.
            Note that there are 1000 classes, so a prediction error of 50\% is
            much better than a random guess.
            The auxiliary head, shown under the model, was observed to have
            a prediction error that was slightly better than the main head.
            This is not necessarily a condition that will hold at the end of training,
            but merely an observation.
            Red is bad (high prediction error) and green/blue is good (low prediction error).
  }
  \label{appfig:inception-model}
\end{figure*}

Here we provide in \figurename ~\ref{appfig:inception-model-sketch}
a sketch of the original Inception v3 model,
and in \figurename ~\ref{appfig:inception-model} we show results from 4 particular moments during training. These are spread over the 2 weeks of training so
that we can get a sense of progression.


As discussed in \sectionname \ref{lcp-sec:dropping-features}, we had to
resort to a technique to limit the number of features used
by the linear classifier probes. In this particular experiment,
we have had the most success by taking 1000 random features for each probe.
This gives certain layers an unfair advantage if they start with 4000 features
and we kept $1000$, whereas in other cases the probe insertion point has $426,320$
features and we keep $1000$. There was no simple ``fair'' solution. That being said,
13 out of the 17 probes have more than $100,000$ features, and 11 of those probes have more
than $200,000$ features, so things were relatively comparable.

\end{document}